\documentclass[11pt,a4paper]{article}
\usepackage[hyperref]{emnlp-ijcnlp-2019}
\usepackage{times}
\usepackage{latexsym}

\usepackage{url}

\aclfinalcopy 

\usepackage{booktabs}
\usepackage{helvet}
\usepackage{courier}
\usepackage{graphicx}
\usepackage{bm}
\usepackage{amssymb}
\usepackage{amsmath}
\usepackage{array}
\usepackage{hyperref}
\usepackage{float}
\usepackage{multirow}
\usepackage{comment}
\usepackage[ruled,vlined]{algorithm2e}

\graphicspath{{./}}

\newcommand{\R}{{\mathbb{R}}}

\newcolumntype{L}[1]{>{\raggedright\let\newline\\\arraybackslash\hspace{0pt}}m{#1}}
\newcolumntype{C}[1]{>{\centering\let\newline\\\arraybackslash\hspace{0pt}}m{#1}}
\newcolumntype{R}[1]{>{\raggedleft\let\newline\\\arraybackslash\hspace{0pt}}m{#1}}

\newcommand{\tabincell}[2]{\begin{tabular}{@{}#1@{}}#2\end{tabular}}

\usepackage{balance}
\usepackage{scrextend}
\deffootnote[1.5em]{0em}{1em}{
\textsuperscript{\thefootnotemark}\space
}

\title{Doc2EDAG: An End-to-End Document-level Framework for \\
Chinese Financial Event Extraction}

\author{
Shun Zheng$^\dagger$\thanks{This work was done during the internship of Shun Zheng at Microsoft Research Asia, Beijing, China.} ,
Wei Cao$^\ddagger$,
Wei Xu$^\dagger$,
Jiang Bian$^\ddagger$ \\
$^\dagger$Institute of Interdisciplinary Information Sciences, Tsinghua University\\
$^\ddagger$Microsoft Research\\
{\tt zhengs14@mails.tsinghua.edu.cn;} \\
{\tt \{Wei.Cao, Jiang.Bian\}@microsoft.com;}\\
{\tt weixu@mail.tsinghua.edu.cn}
}

\date{}

\begin{document}

\maketitle

\begin{abstract}
Most existing event extraction (EE) methods merely extract event arguments within the sentence scope. However, such sentence-level EE methods struggle to handle soaring amounts of documents from emerging applications, such as finance, legislation, health, etc., where event arguments always scatter across different sentences, and even multiple such event mentions frequently co-exist in the same document. To address these challenges, we propose a novel end-to-end model, Doc2EDAG, which can generate an entity-based directed acyclic graph to fulfill the document-level EE (DEE) effectively. Moreover, we reformalize a DEE task with the no-trigger-words design to ease document-level event labeling. To demonstrate the effectiveness of Doc2EDAG, we build a large-scale real-world dataset consisting of Chinese financial announcements with the challenges mentioned above. Extensive experiments with comprehensive analyses illustrate the superiority of Doc2EDAG over state-of-the-art methods.
Data and codes can be found at \url{https://github.com/dolphin-zs/Doc2EDAG}.
\end{abstract}

\section{Introduction}

Event extraction (EE), traditionally modeled as detecting trigger words and extracting corresponding arguments from plain text, plays a vital role in natural language processing since it can produce valuable structured information to facilitate a variety of tasks, such as knowledge base construction, question answering, language understanding, etc.

\begin{figure}[tb]
\centering
\includegraphics[width=0.99\linewidth]{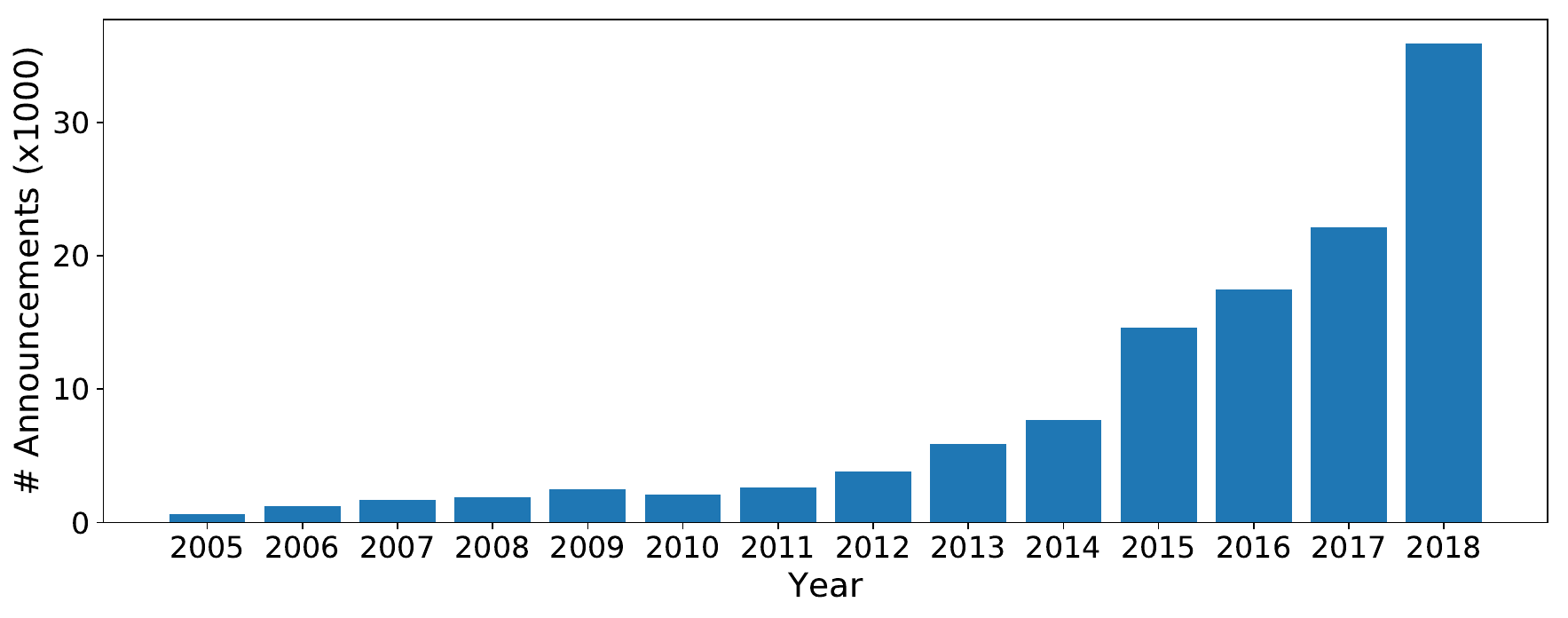}
\caption{The rapid growth of event-related announcements considered in this paper.}
\label{fig:ann_growth}
\end{figure}

In recent years, with the rising trend of digitalization within various domains, such as finance, legislation, health, etc., EE has become an increasingly important accelerator to the development of business in those domains.  
Take the financial domain as an example, continuous economic growth has witnessed exploding volumes of digital financial documents, such as financial announcements in a specific stock market as Figure~\ref{fig:ann_growth} shows, specified as \textit{Chinese financial announcements} (ChFinAnn).
While forming up a gold mine, such large amounts of announcements call EE for assisting people in extracting valuable structured information to sense emerging risks and find profitable opportunities timely.

\begin{figure*}[tb]
\centering
\includegraphics[width=0.99\linewidth]{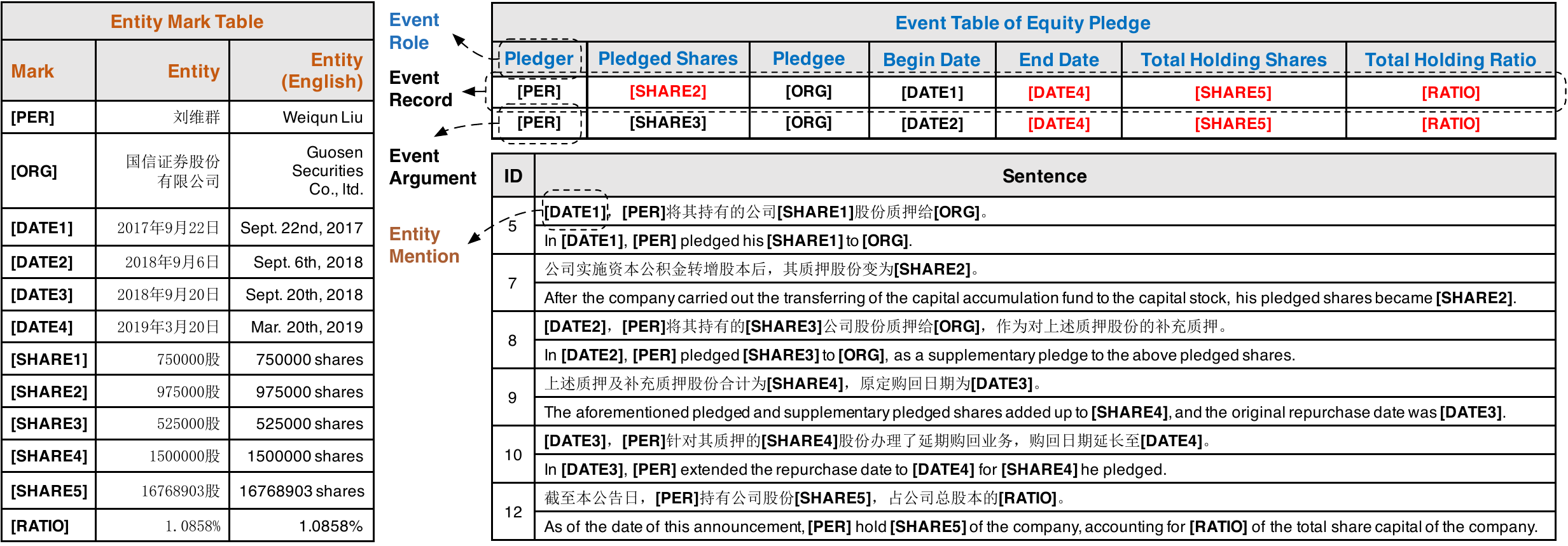}
\caption{A document example with two \textit{Equity Pledge} event records whose arguments scatter across multiple sentences, where we use \textit{ID} to denote the sentence index, substitute entity mentions with corresponding marks, and color event arguments outside the scope of key-event sentences as red.}
\label{fig:pledge_example}
\end{figure*}

Given the necessity of applying EE on the financial domain, the specific characteristics of financial documents as well as those within many other business fields, however, raise two critical challenges to EE, particularly \textit{arguments-scattering} and \textit{multi-event}.  Specifically, the first challenge indicates that arguments of one event record may scatter across multiple sentences of the document, while the other one reflects that a document is likely to contain multiple such event records.
To intuitively illustrate these challenges,
we show a typical ChFinAnn document with two \textit{Equity Pledge} event records in Figure~\ref{fig:pledge_example}.
For the first event, the entity\footnote{In this paper, we use ``entity'' as a general notion that includes named entities, numbers, percentages, etc., for brevity.} ``[SHARE1]'' is the correct \textit{Pledged Shares} at the sentence level (\textit{ID 5}).
However, due to the capital stock increment (\textit{ID 7}), the correct \textit{Pledged Shares} at the document level should be ``[SHARE2]''.
Similarly, ``[DATE3]'' is the correct \textit{End Date} at the sentence level (\textit{ID 9}) but incorrect at the document level (\textit{ID 10}).
Moreover, some summative arguments, such as ``[SHARE5]'' and ``[RATIO]'', are often stated at the end of the document.

Although a great number of efforts~\cite{ahn2006stages,ji2008refining,liao2010using,hong2011using,riedel2011fast,li2013joint,li2014constructing,chen2015event,yang2016joint,nguyen2016joint,liu2017exploiting,sha2018jointly,zhang2018event,nguyen2019one,wang2019adversarial} have been put on EE, most of them are based on ACE 2005\footnote{\url{https://www.ldc.upenn.edu/collaborations/past-projects/ace}},
an expert-annotated benchmark, 
which only tagged event arguments within the sentence scope.
We refer to such task as the \textbf{sentence-level EE} (SEE), which obviously overlooks the \textit{arguments-scattering} challenge.
In contrast, EE on financial documents, such as ChFinAn, requires \textbf{document-level EE} (DEE) when facing \textit{arguments-scattering}, and this challenge gets much harder when coupled with \textit{multi-event}. 

The most recent work, DCFEE~\cite{yang2018dcfee}, attempted to explore DEE on ChFinAnn, by employing \textit{distant supervision} (DS)~\cite{mintz2009distant} to generate EE data and performing a two-stage extraction: 1) a sequence tagging model for SEE, and 2) a key-event-sentence detection model to detect the key-event sentence, coupled with a heuristic strategy that padded missing arguments from surrounding sentences, for DEE.
However, the sequence tagging model for SEE cannot handle multi-event sentences elegantly, and even worse, the context-agnostic arguments-completion strategy fails to address the arguments-scattering challenge effectively.

In this paper, we propose a novel end-to-end model, Doc2EDAG, to address the unique challenges of DEE.
The key idea of Doc2EDAG is to transform the event table into an \textit{entity-based directed acyclic graph} (EDAG).
The EDAG format can transform the hard table-filling task into several sequential path-expanding sub-tasks that are more tractable.
To support the EDAG generation efficiently, Doc2EDAG encodes entities with document-level contexts and designs a memory mechanism for path expanding.
Moreover, to ease the DS-based document-level event labeling,
we propose a novel DEE formalization that removes the trigger-words labeling and regards DEE as directly filling event tables based on a document.
This no-trigger-words design does not rely on any predefined trigger-words set or heuristic to filter multiple trigger candidates, and still perfectly matches the ultimate goal of DEE, mapping a document to underlying event tables.

To evaluate the effectiveness of our proposed Doc2EDAG, we conduct experiments on a real-world dataset, consisting of large scales of financial announcements. In contrast to the dataset used by DCFEE where
97\%\footnote{Estimated by their Table 1 as $\frac{2*\text{NO.ANN}-\text{NO.POS}}{\text{NO.ANN}}$.}
documents just contained one event record,
our data collection is ten times larger where about 30\% documents include multiple event records. 
Extensive experiments demonstrate that Doc2EDAG can significantly outperform state-of-the-art methods when facing DEE-specific challenges.

In summary, our contributions include:
\begin{itemize}
    \item We propose a novel model, Doc2EDAG, which can directly generate event tables based on a document, to address unique challenges of DEE effectively.
    \item We reformalize a DEE task without trigger words to ease the DS-based document-level event labeling.
    \item We build a large-scale real-world dataset for DEE with the unique challenges of \textit{arguments-scattering} and \textit{multi-event}, the extensive experiments on which demonstrate the superiority of Doc2EDAG.
\end{itemize}

Note that though we focus on ChFinAnn data in this work, we tackle those DEE-specific challenges without any domain-specific assumption.
Therefore, our general labeling and modeling strategies can directly benefit many other business domains with similar challenges, such as criminal facts and judgments extraction from legal documents, disease symptoms and doctor instructions identification from medical reports, etc.


\section{Related Work}

Recent development on information extraction has been advancing in building the joint model that can extract entities and identify structures (relations or events) among them simultaneously.
For instance,~\cite{ren2017cotype,Zheng2017JointEO,zeng2018extracting,wang2018joint} focused on jointly extracting entities and inter-entity relations.
In the meantime, the same to the focus of this paper, a few studies aimed at designing joint models for the entity and event extraction,
such as handcrafted-feature-based~\cite{li2014constructing,yang2016joint,judea2016incremental} and neural-network-based~\cite{zhang2018event,nguyen2019one} models.
Nevertheless, these models did not present how to handle argument candidates beyond the sentence scope.
\cite{yang2016joint} claimed to handle event-argument relations across sentences with the prerequisite of well-defined features, which, unfortunately, is nontrivial.

In addition to the modeling challenge, another big obstacle for democratizing EE is the lack of training data due to the enormous cost to obtain expert annotations.
To address this problem, some researches attempted to adapt distant supervision (DS) to the EE setting, since DS has shown promising results by employing knowledge bases to automatically generate training data for relation extraction~\cite{mintz2009distant}.
However, the vanilla EE required the trigger words that were absent on factual knowledge bases.
Therefore,~\cite{chen2017automatically,yang2018dcfee} employed either linguistic resources or predefined dictionaries for trigger-words labeling.
On the other hand, another recent work~\cite{zeng2018scale} showed that directly labeling event arguments without trigger words was also feasible.
However, they only considered the SEE setting and their methods cannot be directly extended to the DEE setting, which is the main focus of this work.

Traditionally, when applying DS to relation extraction, researchers put huge efforts into alleviating labeling noises~\cite{riedel2010modeling,lin2016neural,feng2018reinforcement,zheng2019diagnre}.
In contrast, this work shows that combining DS with some simple constraints can obtain pretty good labeling quality for DEE, where the reasons are two folds:
1) both the knowledge base and text documents are from the same domain;
2) an event record usually contains multiple arguments, while a common relational fact only covers two entities.


\section{Preliminaries}

We first clarify several key notions:
1) \textbf{entity mention}:
an entity mention is a text span that refers to an entity object;
2) \textbf{event role}:
an event role corresponds to a predefined field of the event table;
3) \textbf{event argument}:
an event argument is an entity that plays a specific event role;
4) \textbf{event record}:
an event record corresponds to an entry of the event table and contains several arguments with required roles.
For example, Figure~\ref{fig:pledge_example} shows two event records, where the entity ``[PER]'' is an event argument with the \textit{Pledger} role.

To better elaborate and evaluate our proposed approach, we leverage the ChFinAnn data in this paper.
ChFinAnn documents contain firsthand official disclosures of listed companies in the Chinese stock market and have hundreds of types, such as annual reports and earnings estimates.
While in this work, we focus on those event-related ones that are frequent, influential, and mainly expressed by the natural language.

\section{Document-level Event Labeling}
\label{sec:doc_event_labeling}

As a prerequisite to DEE, we first conduct the DS-based event labeling at the document level.
More specifically, we map tabular records from an event knowledge base to document text and regard well-matched records as events expressed by that document.
Moreover, we adopt a no-trigger-words design and reformalize a novel DEE task accordingly to enable end-to-end model designs.

\paragraph{Event Labeling.}
To ensure the labeling quality, we set two constraints for matched records: 1)~arguments of predefined key event roles must exist (non-key ones can be empty)
and~2)~the number of matched arguments should be higher than a certain threshold.
Configurations of these constraints are event-specific, and in practice, we can tune them to directly ensure the labeling quality at the document level.
We regard records that meet these two constraints as the well-matched ones, which serve as distantly supervised ground truths.
In addition to labeling event records, we assign roles of arguments to matched tokens as token-level entity tags.
Note that we do not label trigger words explicitly.
Besides not affecting the DEE functionality, an extra benefit of such no-trigger-words design is a much easier DS-based labeling that does not rely on predefined trigger-words dictionaries or manually curated heuristics to filter multiple potential trigger words.

\paragraph{DEE Task Without Trigger Words.}
We reformalize a novel task for DEE as directly filling event tables based on a document, which generally requires three sub-tasks:
1)~\textbf{entity extraction}, extracting entity mentions as argument candidates,
2)~\textbf{event detection}, judging a document to be triggered or not for each event type,
and 3)~\textbf{event table filling}, filling arguments into the table of triggered events.
This novel DEE task is much different from the vanilla SEE with trigger words but is consistent with the above simplified DS-based event labeling.


\section{Doc2EDAG}
\begin{figure}[t]
\centering
\includegraphics[width=0.99\linewidth]{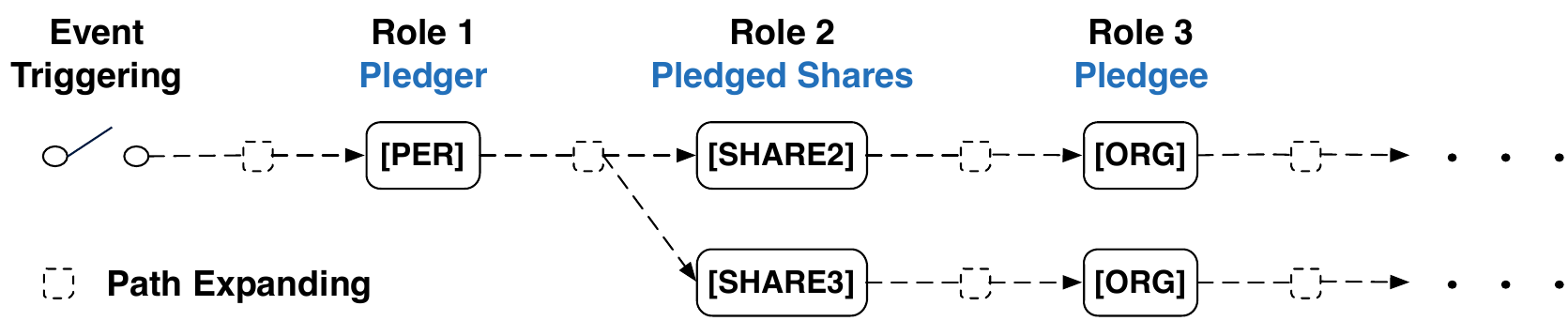}
\caption{An EDAG generation example that starts from event triggering and expands sequentially following the predefined order of event roles.}
\label{fig:edag_example}
\end{figure}

\begin{figure*}[ht]
\centering
\includegraphics[width=0.99\linewidth]{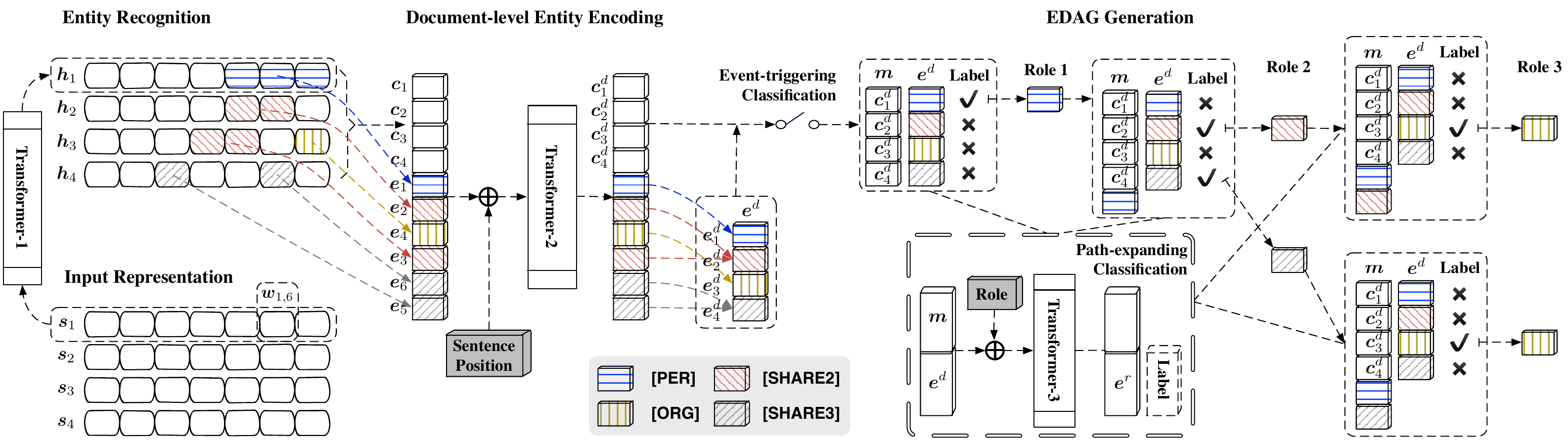}
\caption{The overall workflow of Doc2EDAG, where we follow the example in Figure~\ref{fig:pledge_example} and the EDAG structure in Figure~\ref{fig:edag_example}, and use stripes to differentiate different entities (note that the number of input tokens and entity positions are imaginary, which do not match previous ones strictly, and here we only include the first three event roles and associated entities for brevity).}
\label{fig:architecture}
\end{figure*}

The key idea of Doc2EDAG is to transform tabular event records into an EDAG and let the model learn to generate this EDAG based on document-level contexts.
Following the example in Figure~\ref{fig:pledge_example},
Figure~\ref{fig:edag_example} typically depicts an EDAG generation process and Figure~\ref{fig:architecture} presents the overall workflow of Doc2EDAG, which consists of two key stages: document-level entity encoding (Section~\ref{sec:doc_ent_enc}) and EDAG generation (Section \ref{sec:dag_gen}).
Before elaborating each of them in this section, we first describe two preconditioned modules: input representation and entity recognition.

\paragraph{Input Representation.}
In this paper, we denote a document as a sequence of sentences.
Formally, after looking up the token embedding table $\bm{V} \in \R^{d_w \times |V|}$,
we denote a document $\bm{d}$ as a sentence sequence $[\bm{s_1};\bm{s_2};\cdots;\bm{s_{N_s}}]$ and each sentence $\bm{s_i} \in \R^{d_w \times N_w}$ is composed of a sequence of token embeddings as $[\bm{w_{i,1}},\bm{w_{i,2}},\cdots,\bm{w_{i,N_w}}]$,
where $|V|$ is the vocabulary size,
$N_s$ and $N_w$ are the maximum lengths of the sentence sequence and the token sequence, respectively, 
and $\bm{w_{i,j}} \in \R^{d_w}$ is the embedding of $j^{th}$ token in $i^{th}$ sentence with the embedding size $d_w$.

\paragraph{Entity Recognition.}
Entity recognition is a typical sequence tagging task.
We conduct this task at the sentence level and follow a classic method, BI-LSTM-CRF~\cite{huang2015bidirectional}, that first encodes the token sequence and then adds a conditional random field (CRF) layer to facilitate the sequence tagging.
The only difference is that we employ the Transformer~\cite{vaswani2017attention} instead of the original encoder, LSTM~\cite{hochreiter1997long}.
Transformer encodes a sequence of embeddings by the multi-headed self-attention mechanism to exchange contextual information among them.
Due to the superior performance of the Transformer, we employ it as a primary context encoder in this work and name the Transformer module used in this stage as Transformer-1.
Formally, for each sentence tensor $\bm{s_i} \in \R^{d_w \times N_w}$,
we get the encoded one as $\bm{h_i} = \text{Transformer-1}(\bm{s_i})$,
where $\bm{h_i} \in \R^{d_w \times N_w}$ shares the same embedding size $d_w$ and sequence length $N_w$.
During training, we employ roles of matched arguments as entity labels with the classic \texttt{BIO} (Begin, Inside, Other) scheme and wrap $\bm{h_i}$ with a CRF layer to get the entity-recognition loss $L_{er}$.
As for the inference, we use the Viterbi decoding to get the best tagging sequence.

\subsection{Document-level Entity Encoding}
\label{sec:doc_ent_enc}
To address the arguments-scattering challenge efficiently,
it is indispensable to leverage global contexts to better identify whether an entity plays a specific event role.
Consequently, we utilize document-level entity encoding to encode extracted entity mentions with such contexts and produce an embedding of size $d_w$ for each entity mention with a distinct surface name.

\paragraph{Entity \& Sentence Embedding.}
Since an entity mention usually covers multiple tokens with a variable length,
we first obtain a fixed-sized embedding for each entity mention by conducting a max-pooling operation over its covered token embeddings.
For example, given $l^{th}$ entity mention covering $j^{th}$ to $k^{th}$ tokens of $i^{th}$ sentence, we conduct the max-pooling over $[\bm{h_{i,j}},\cdots,\bm{h_{i,k}}]$ to get the entity mention embedding $\bm{e}_l \in \R^{d_w}$.
For each sentence $\bm{s_i}$, we also take the max-pooling operation over the encoded token sequence $[\bm{h_{i,1}},\cdots,\bm{h_{i,N_w}}]$ to obtain a single sentence embedding $\bm{c_i} \in \R^{d_w}$.
After these operations, both the mention and the sentence embeddings share the same embedding size $d_w$.

\paragraph{Document-level Encoding.}
Though we get embeddings for all sentences and entity mentions, these embeddings only encode local contexts within the sentence scope.
To enable the awareness of document-level contexts, we employ the second Transformer module, Transformer-2, to facilitate the information exchange between all entity mentions and sentences.
Before feeding them into Transformer-2, we add them with sentence position embeddings to inform the sentence order.
After the Transformer encoding, we utilize the max-pooling operation again to merge multiple mention embeddings with the same entity surface name into a single embedding.
Formally, after this stage, we obtain document-level context-aware entity mention and sentence embeddings as
$\bm{e^d} = [\bm{e^d_1},\cdots,\bm{e^d_{N_e}}]$ and $\bm{c^d} = [\bm{c^d_1},\cdots,\bm{c^d_{N_s}}]$, respectively,
where $N_e$ is the number of distinct entity surface names.
These aggregated embeddings serve the next stage to fill event tables directly.

\subsection{EDAG Generation}
\label{sec:dag_gen}
After the document-level entity encoding stage, we can obtain the document embedding $\bm{t} \in \R^{d_w}$ by operating the max-pooling over the sentence tensor $\bm{c^d} \in \R^{d_w\times N_s}$ and stack a linear classifier over $\bm{t}$ to conduct the event-triggering classification for each event type.
Next, for each triggered event type, we learn to generate an EDAG.

\paragraph{EDAG Building.}
Before the model training, we need to build the EDAG from tabular event records.
For each event type, we first manually define an event role order.
Then, we transform each event record into a linked list of arguments following this order, where each argument node is either an entity or a special empty argument \texttt{NA}.
Finally, we merge these linked lists into an EDAG by sharing the same prefix path.
Since every complete path of the EDAG corresponds to one row of the event table, recovering the table format from a given EDAG is simple.

\paragraph{Task Decomposition.}
The EDAG format aims to simplify the hard table-filling task into several tractable path-expanding sub-tasks.
Then, a natural question is how the task decomposition works, which can be answered by the following EDAG recovering procedure.
Assume the event triggering as the starting node (the initial EDAG), there comes a series of path-expanding sub-tasks following a predefined event role order. 
When considering a certain role, for every leaf node of the current EDAG, there is a path-expanding sub-task that decides which entities to be expanded.
For each entity to be expanded, we create a new node of that entity for the current role and expand the path by connecting the current leaf node to the new entity node.
If no entity is valid for expanding, we create a special \texttt{NA} node.
When all sub-tasks for the current role finish, we move to the next role and repeat until the last.
In this work, we leverage the above logic to recover the EDAG from path-expanding predictions at inference and to set associated labels for each sub-task when training.

\paragraph{Memory.}
To better fulfill each path-expanding sub-task, it is crucial to know entities already contained by the path.
Hence, we design a memory mechanism that initializes a memory tensor $\bm{m}$ with the sentence tensor $\bm{c^d}$ at the beginning and updates $\bm{m}$ when expanding the path by appending either the associated entity embedding or the zero-padded one for the \texttt{NA} argument.
With this design, each sub-task can own a distinct memory tensor, corresponding to the unique path history.

\paragraph{Path Expanding.}
For each path-expanding sub-task, we formalize it as a collection of multiple binary classification problems, that is predicting \textit{expanding} (1) or \textit{not} (0) for all entities.
To enable the awareness of the current path state, history contexts and the current event role,
we first concatenate the memory tensor $\bm{m}$ and the entity tensor $\bm{e^d}$,
then add them with a trainable event-role-indicator embedding,
and encode them with the third Transformer module, Transformer-3, to facilitate the context-aware reasoning.
Finally, we extract the enriched entity tensor $\bm{e^r}$ from outputs of Transformer-3 and stack a linear classifier over $\bm{e^r}$ to conduct the path-expanding classification.

\paragraph{Optimization.}
For the event-triggering classification, we calculate the cross-entropy loss $L_{tr}$.
During the EDAG generation, we calculate a cross-entropy loss for each path-expanding sub-task, and sum these losses as the final EDAG-generation loss $L_{dag}$.
Finally, we sum $L_{tr}$, $L_{dag}$ and the entity-recognition loss $L_{er}$ together as the final loss, $L_{all} = \lambda_1 L_{er} + \lambda_2 L_{tr} + \lambda_3 L_{dag}$, where $\lambda_1$, $\lambda_2$ and $\lambda_3$ are hyper-parameters.

\paragraph{Inference.}
Given a document, Doc2EDAG first recognizes entity mentions from sentences, 
then encodes them with document-level contexts, 
and finally generates an EDAG for each triggered event type by conducting a series of path-expanding sub-tasks.

\paragraph{Practical Tips.}
During training, we can utilize both ground-truth entity tokens and the given EDAG structure.
While at inference, we need to first identify entities and then expand paths sequentially based on embeddings of those entities to recover the EDAG.
This gap between training and inference can cause severe error-propagation problems.
To mitigate such problems, we utilize the scheduled sampling~\cite{bengio2015scheduled} to gradually switch the inputs of document-level entity encoding from ground-truth entity mentions to model recognized ones.
Moreover, for path-expanding classifications, false positives are more harmful than false negatives, because the former can cause a completely wrong path. Accordingly, we can set $\gamma (> 1)$ as the negative class weight of the associated cross-entropy loss.


\section{Experiments}
In this section, we present thorough empirical studies to answer the following questions:
1)~to what extent can Doc2EDAG improve over state-of-the-art methods when facing DEE-specific challenges?
2)~how do different models behave when facing both \textit{arguments-scattering} and \textit{multi-event} challenges?
3)~how important are various components of Doc2EDAG?

\subsection{Experimental Setup}

\paragraph{Data Collection with Event Labeling.}
We utilize ten years (2008-2018) ChFinAnn\footnote{Crawling from \url{http://www.cninfo.com.cn/new/index}} documents and human-summarized event knowledge bases to conduct the DS-based event labeling.
We focus on five event types: \textit{Equity Freeze} (EF), \textit{Equity Repurchase} (ER),
\textit{Equity Underweight} (EU), \textit{Equity Overweight} (EO) and \textit{Equity Pledge} (EP),
which belong to major events required to be disclosed by the regulator and may have a huge impact on the company value.
To ensure the labeling quality, we set constraints for matched document-record pairs as Section~\ref{sec:doc_event_labeling} describes.
Moreover, we directly use the character tokenization to avoid error propagations from Chinese word segmentation tools.

\begin{table}[tb]\small
\centering
\addtolength{\tabcolsep}{-1.5pt}
\begin{tabular}{c r r r r r}
\toprule
\textbf{Event} & \textbf{\#Train} & \textbf{\#Dev} & \textbf{\#Test} & \textbf{\#Total} & \textbf{MER (\%)} \\
\midrule
 EF & $806$    & $186$   & $204$   & $1,196$  & $32.0$  \\
 ER & $1,862$  & $297$   & $282$   & $3,677$  & $16.1$  \\
 EU & $5,268$  & $677$   & $346$   & $5,847$  & $24.3$  \\
 EO & $5,101$  & $570$   & $1,138$ & $6,017$  & $28.0$  \\
 EP & $12,857$ & $1,491$ & $1,254$ & $15,602$ & $35.4$  \\ 
\midrule
All & $25,632$ & $3,204$ & $3,204$ & $32,040$ & $29.0$ \\
\bottomrule
\end{tabular}
\caption{Dataset statistics about the number of documents for the train (\#Train), development (\#Dev) and test (\#Test), the number (\#Total) and the multi-event ratio (MER) of all documents.}
\label{tab:dataset}
\end{table}

\begin{table}[tb]\small
\centering
\addtolength{\tabcolsep}{-1.5pt}
\begin{tabular}{c c c c}
\toprule
\textbf{Precision} & \textbf{Recall} & \textbf{F1} & \textbf{MER (\%)} \\
\midrule
98.8 & 89.7 & 94.0 & 31.0
\\
\bottomrule
\end{tabular}
\caption{The quality of the DS-based event labeling evaluated on $100$ manually annotated documents (randomly select $20$ for each event type).}
\label{tab:label_quality}
\end{table}

Finally, we obtain $32,040$ documents in total,
and this number is ten times larger than $2,976$ of DCFEE and about $53$ times larger than $599$ of ACE 2005.
We divide these documents into train, development, and test set with the proportion of $8:1:1$ based on the time order.
In Table~\ref{tab:dataset}, we show the number of documents and the multi-event ratio (MER) for each event type on this dataset.
Note that a few documents may contain multiple event types at the same time.

\paragraph{Data Quality.}
To verify the quality of DS-based event labeling,
we randomly select $100$ documents and manually annotate them.
By regarding DS-generated event tables as the prediction and human-annotated ones as the ground-truth,
we evaluate the labeling quality based on the metric introduced below.
Table~\ref{tab:label_quality} shows this approximate evaluation,
and we can observe that DS-generated data are pretty good, achieving high precision and acceptable recall.
In later experiments, we directly employ the automatically generated test set for evaluation due to its much broad coverage.

\paragraph{Evaluation Metric.}
The ultimate goal of DEE is to fill event tables with correct arguments for each role.
Therefore, we evaluate DEE by directly comparing the predicted event table with the ground-truth one for each event type.
Specifically, for each document and each event type, we pick one predicted record and one most similar ground-truth record (at least one of them is non-empty) from associated event tables without replacement to calculate event-role-specific true positive, false positive and false negative statistics until no record left.
After aggregating these statistics among all evaluated documents, we can calculate role-level precision, recall, and F1 scores (all reported in percentage format).
As an event type often includes multiple roles, we calculate micro-averaged role-level scores as the final event-level metric that reflects the ability of end-to-end DEE directly.

\paragraph{Hyper-parameter Setting.}
For the input, we set the maximum number of sentences and the maximum sentence length as $64$ and $128$, respectively.
During training, we set $\lambda_1=0.05$, $\lambda_2=\lambda_3=0.95$ and $\gamma=3$.
We employ the Adam~\cite{kingma2014adam} optimizer with the learning rate $1e^{-4}$,
train for at most $100$ epochs and pick the best epoch by the validation score on the development set.
Besides, we leverage the decreasing order of the non-empty argument ratio as the event role order required by Doc2EDAG, because more informative entities in the path history can better facilitate later path-expanding classifications.

Note that, due to the space limit, we leave other detailed hyper-parameters, model structures, data preprocessing configurations, event type specifications and pseudo codes for EDAG generation to the appendix.

\begin{table*}[tb]\small
\centering
\addtolength{\tabcolsep}{-1pt}
\begin{tabular}{l | ccc | ccc | ccc | ccc | ccc }
\toprule
\multirow{2}{*}{\textbf{Model}}
& \multicolumn{3}{c|}{\textbf{EF}}
& \multicolumn{3}{c|}{\textbf{ER}}
& \multicolumn{3}{c|}{\textbf{EU}}
& \multicolumn{3}{c|}{\textbf{EO}}
& \multicolumn{3}{c}{\textbf{EP}}
\\
& \textbf{P.} & \textbf{R.} & \textbf{F1}
& \textbf{P.} & \textbf{R.} & \textbf{F1}
& \textbf{P.} & \textbf{R.} & \textbf{F1}
& \textbf{P.} & \textbf{R.} & \textbf{F1}
& \textbf{P.} & \textbf{R.} & \textbf{F1}
\\
\midrule
DCFEE-O  & 66.0 & 41.6 & 51.1      & 84.5 & 81.8 & 83.1      & 62.7 & 35.4 & 45.3      & 51.4 & 42.6 & 46.6      & 64.3 & 63.6 & 63.9      
\\
DCFEE-M & 51.8 & 40.7 & 45.6 	 & 83.7 & 78.0 & 80.8 	 & 49.5 & 39.9 & 44.2 	 & 42.5 & 47.5 & 44.9 	 & 59.8 & 66.4 & 62.9 
\\
\midrule
GreedyDec  
& \textbf{79.5} & 46.8 & 58.9      & 83.3 & 74.9 & 78.9      & 68.7 & 40.8 & 51.2      & 69.7 & 40.6 & 51.3      & \textbf{85.7} & 48.7 & 62.1
\\
Doc2EDAG  
& 77.1 & \textbf{64.5} & \textbf{70.2}      & \textbf{91.3} & \textbf{83.6} & \textbf{87.3}      & \textbf{80.2} & \textbf{65.0} & \textbf{71.8}      & \textbf{82.1} & \textbf{69.0} & \textbf{75.0}      & 80.0 & \textbf{74.8} & \textbf{77.3}
\\
\bottomrule
\end{tabular}
\caption{Overall event-level precision (P.), recall (R.) and F1 scores evaluated on the test set.}
\label{tab:main_result}
\end{table*}

\begin{table*}[tb]\small
\centering
\addtolength{\tabcolsep}{-1pt}
\begin{tabular}{l | cc | cc | cc | cc | cc | ccc}
\toprule
\multirow{2}{*}{\textbf{Model}}
& \multicolumn{2}{c|}{\textbf{EF}}
& \multicolumn{2}{c|}{\textbf{ER}}
& \multicolumn{2}{c|}{\textbf{EU}}
& \multicolumn{2}{c|}{\textbf{EO}}
& \multicolumn{2}{c|}{\textbf{EP}}
& \multicolumn{3}{c}{\textbf{Avg.}}
\\
& \textbf{S.} & \textbf{M.}
& \textbf{S.} & \textbf{M.}
& \textbf{S.} & \textbf{M.}
& \textbf{S.} & \textbf{M.}
& \textbf{S.} & \textbf{M.}
& \textbf{S.} & \textbf{M.} & \textbf{S. \& M.}
\\
\midrule
DCFEE-O  
& 56.0 & 46.5          & 86.7 & 54.1          & 48.5 & 41.2          & 47.7 & 45.2          & 68.4 & 61.1 & 61.5 & 49.6 & 58.0
\\
DCFEE-M  
& 48.4 & 43.1     	 & 83.8 & 53.4     	 & 48.1 & 39.6     	 & 47.1 & 42.0     	 & 67.0 & 60.6  & 58.9 & 47.7 & 55.7
\\
\midrule
GreedyDec  
& 75.9 & 40.8          & 81.7 & 49.8          & 62.2 & 34.6          & 65.7 & 29.4      & \textbf{88.5} & 42.3  & 74.8 & 39.4 & 60.5
\\
Doc2EDAG 
& \textbf{80.0} & \textbf{61.3}          & \textbf{89.4} & \textbf{68.4}          & \textbf{77.4} & \textbf{64.6}          & \textbf{79.4} & \textbf{69.5}          & 85.5 & \textbf{72.5}  & \textbf{82.3} & \textbf{67.3} &  \textbf{76.3}
\\
\bottomrule
\end{tabular}
\caption{F1 scores for all event types and the averaged ones (Avg.) on single-event (S.) and multi-event (M.) sets.}
\label{tab:single_multi}
\end{table*}

\begin{table}[tb]\small
\centering
\addtolength{\tabcolsep}{-1.55pt}
\begin{tabular}{l | r | r | r | r | r | r }
\toprule
\textbf{Model} & \textbf{EF} & \textbf{ER} & \textbf{EU} & \textbf{EO} & \textbf{EP} & \textbf{Avg.}
\\
\midrule
Doc2EDAG 
& 70.2  & 87.3 & 71.8  & 75.0  & 77.3  & 76.3
\\
\midrule
-PathMem  
& -11.2 & -0.2 & -10.1 & -16.3 & -10.9 & -9.7
\\
-SchSamp
& -5.3 & -4.8 & -5.3 & -6.6 & -3.0 & -5.0
\\
-DocEnc  
& -4.7  & -1.5 & -1.6  & -1.1  & -1.5  & -2.1
\\
-NegCW
& -1.4 & -0.4 & -0.7 & -1.3 & -0.4 & -0.8
\\
\bottomrule
\end{tabular}
\caption{Performance differences of Doc2EDAG variants for all event types and the averaged ones (Avg.).}
\label{tab:ablation_study}
\end{table}

\subsection{Performance Comparisons}
\paragraph{Baselines.}
As discussed in the related work, the state-of-the-art method applicable to our setting is DCFEE.
We follow the implementation described in~\cite{yang2018dcfee},
but they did not illustrate how to handle multi-event sentences with just a sequence tagging model.
Thus, we develop two versions, \textit{DCFEE-O} and \textit{DCFEE-M},
where \textit{DCFEE-O} only produces one event record from one key-event sentence,
while \textit{DCFEE-M} tries to get multiple possible argument combinations by the closest relative distance from the key-event sentence.
To be fair, the SEE stages of both versions share the same neural architecture as the entity recognition part of Doc2EDAG.
Besides, we further employ a simple decoding baseline of Doc2EDAG, \textit{GreedyDec},
that only fills one event table entry greedily by using recognized entity roles to verify the necessity of end-to-end modeling.

\paragraph{Main Results.}
As Table~\ref{tab:main_result} shows, Doc2EDAG achieves significant improvements over all baselines for all event types.
Specifically, Doc2EDAG improves $19.1$, $4.2$, $26.5$, $28.4$ and $13.4$ F1 scores over DCFEE-O, the best baseline, on EF, ER, EU, EO and EP events, respectively.
These vast improvements mainly owe to the document-level end-to-end modeling of Doc2EDAG.
Moreover, since we work on automatically generated data,
the direct document-level supervision can be more robust than the extra sentence-level supervision used in DCFEE,
which assumes the sentence containing most event arguments as the key-event one.
This assumption does not work well on some event types,
such as EF, EU and EO,
on which DCFEE-O is even inferior to the most straightforward baseline, GreedyDec.
Besides, DCFEE-O achieves better results than DCFEE-M, which demonstrates that naively guessing multiple events from the key-event sentence cannot work well.
By comparing Doc2EDAG with GreedyDec that owns high precision but low recall,
we can clearly see the benefit of document-level end-to-end modeling.

\paragraph{Single-Event vs. Multi-Event.}
We divide the test set into a single-event set, containing documents with just one event record, and a multi-event set, containing others, to show the extreme difficulty when \textit{arguments-scattering} meets \textit{multi-event}.
Table~\ref{tab:single_multi} shows F1 scores for different scenarios.
Although Doc2EDAG still maintains the highest extraction performance for all cases,
the multi-event set is extremely challenging as the extraction performance of all models drops significantly.
Especially, GreedyDec, with no mechanism for the \textit{multi-event} challenge, decreases most drastically.
DCFEE-O decreases less, but is still far away from Doc2EDAG.
On the multi-event set, Doc2EDAG increases by $17.7$ F1 scores over DCFEE-O, the best baseline, on average.

\paragraph{Ablation Tests.}
To demonstrate key designs of Doc2EDAG,
we conduct ablation tests by evaluating four variants:
1)~\textit{-PathMem}, removing the memory mechanism used during the EDAG generation,
2)~\textit{-SchSamp}, dropping the scheduled sampling strategy during training,
3)~\textit{-DocEnc}, removing the Transformer module used for document-level entity encoding,
and 4)~\textit{-NegCW}, keeping the negative class weight as $1$ when doing path-expanding classifications.
From Table~\ref{tab:ablation_study}, we can observe that
1)~the memory mechanism is of prime importance, as removing it can result in the most drastic performance declines, over $10$ F1 scores on four event types except for the ER type whose MER is very low on the test set;
2)~the scheduled sampling strategy that alleviates the mismatch of entity candidates for event table filling between training and inference also contributes greatly, improving by $5$ F1 scores on average;
3)~the document-level entity encoding that enhances global entity representations contributes $2.1$ F1 scores on average;
4)~the larger negative class weight to penalize false positive path expanding can also make slight but stable contributions for all event types.

\paragraph{Case Studies.}
Let us follow the example in Figure~\ref{fig:pledge_example}, Doc2EDAG can successfully recover the correct EDAG, while DCFEE inevitably makes many mistakes even with a perfect SEE model, as discussed in the introduction.
Due to the space limit, we leave another three fine-grained case studies to the appendix.


\section{Conclusion and Future Work}
Towards the end-to-end modeling for DEE, we propose a novel model, Doc2EDAG, associated with a novel task formalization without trigger words to ease DS-based labeling.
To validate the effectiveness of the proposed approach, we build a large-scale real-world dataset in the financial domain and conduct extensive empirical studies.
Notably, without any domain-specific assumption, our general labeling and modeling strategies can benefit practitioners in other domains directly.

As this work shows promising results for the end-to-end DEE, expanding the inputs of Doc2EDAG from pure text sequences to richly formatted ones~\cite{wu2018fonduer} is appealing, and we leave it as future work to explore.

\section*{Acknowledgements}
This work is supported in part by the National Natural Science Foundation of China (NSFC) Grant 61532001 and the Zhongguancun Haihua Institute for Frontier Information Technology.

\balance
\bibliographystyle{acl_natbib}
\bibliography{ref_short}

\newpage
\appendix

\section{Appendix}
\label{sec:appendix}

In the appendix, we incorporate the following details that are omitted in the main body due to the space limit.
\begin{itemize}
    \item Section~\ref{sec:event} presents event type specifications and corresponding preprocessing details.
    \item Section~\ref{sec:hyper_para} includes the hyper-parameter setting that we use to run experiments.
    \item Section~\ref{sec:pseudo_codes} provides pseudo codes to facilitate understanding of the EDAG generation.
    \item Section~\ref{sec:add_eval_res} complements additional evaluation results for entity extraction and event triggering.
    \item Section~\ref{sec:case} studies another three sophisticated cases to intuitively illustrate necessities and advantages of end-to-end modeling.
\end{itemize}

\subsection{Event Type Specifications}
\label{sec:event}
Table~\ref{table:event} shows detailed illustrations for event types used in our paper,
where we mark some key roles that should be non-empty when conducting the document-level event labeling.
In addition to requiring non-empty key roles, we empirically set the minimum number of matched roles for EF, ER, EU, EO and EP events as 5, 4, 4, 4 and 5, respectively.
Though we set these constraints empirically when processing our data, practitioners of other domains can adjust these configurations freely to fulfill the task-specific requirements by making a desirable trade-off between precision and recall.

\begin{table*}[htb]\small
\centering
\begin{tabular}{ c | l | l}
\toprule
\textbf{Event Type} & \textbf{Event Role} & \textbf{Detailed Explanations} \\
\midrule
\multirow{8}{*}{\tabincell{c}{Equity \\ Freeze \\ (EF)}}
& \textbf{Equity Holder} (key) & the equity holder whose shares are froze \\
& \textbf{Froze Shares} (key) & the number of shares being froze \\
& \textbf{Legal Institution} (key) & the legal institution that executes this freeze \\
& \textbf{Start Date} &  the start date of this freeze \\
& \textbf{End Date} & the end date of this freeze \\
& \textbf{Unfroze Date} & the date in which these shares are unfroze \\
& \textbf{Total Holding Shares} & the total number of shares being hold at disclosing time \\
& \textbf{Total Holding Ratio} & the total ratio of shares being hold at disclosing time \\
\midrule
\multirow{6}{*}{\tabincell{c}{Equity \\ Repurchase \\ (ER)}}
& \textbf{Company Name} (key) & the name of the company \\
& \textbf{Highest Trading Price} & the highest trading price \\ 
& \textbf{Lowest Trading Price} & the lowest trading price \\
& \textbf{Closing Date} & the closing date of this disclosed repurchase \\
& \textbf{Repurchased Shares} & the number of shares being repurchased before the closing date \\
& \textbf{Repurchase Amount} & the repurchase amount before the closing date \\
\midrule
\multirow{6}{*}{\tabincell{c}{Equity \\ Underweight \\ (EU)}}
& \textbf{Equity Holder} (key) & the equity holder who conducts this underweight \\
& \textbf{Traded Shares} (key) & the number of shares being traded \\
& \textbf{Start Date} & the start date of this underweight \\
& \textbf{End Date} & the end date of this underweight \\
& \textbf{Average Price} & the average price during this underweight \\
& \textbf{Later Holding Shares} & the number of shares being hold after this underweight \\
\midrule
\multirow{6}{*}{\tabincell{c}{Equity \\ Overweight \\ (EO)}}
& \textbf{Equity Holder} (key) & the equity holder who conducts this overweight \\
& \textbf{Traded Shares} (key) & the number of shares being traded \\
& \textbf{Start Date} & the start date of this overweight \\
& \textbf{End Date} & the end date of this overweight \\
& \textbf{Average Price} & the average price during this overweight \\
& \textbf{Later Holding Shares} & the number of shares being hold after this overweight \\
\midrule
\multirow{9}{*}{\tabincell{c}{Equity \\ Pledge \\ (EP)}}
& \textbf{Pledger} (key) & the equity holder who pledges some shares to an institution \\
& \textbf{Pledged Shares} (key) & the number of shares being pledged \\
& \textbf{Pledgee} (key) & the institution who accepts the pledged shares \\
& \textbf{Start Date} & the start date of this pledge \\
& \textbf{End Date} & the end date of this pledge \\
& \textbf{Released Date} & the date in which these pledged shares are released  \\
& \textbf{Total Pledged Shares} & the total number of shares being pledged at disclosing time \\
& \textbf{Total Holding Shares} & the total number of shares being hold at disclosing time \\
& \textbf{Total Holding Ratio} & the total ratio of shares being hold at disclosing time \\
\bottomrule
\end{tabular}
\caption{Event type specifications.}
\label{table:event}
\end{table*}

\subsection{Hyper-parameters Setting}
\label{sec:hyper_para}

We summarize all hyper-parameters in Table~\ref{table:hyper_para} for better reproducibility.

Moreover, when training models, we follow the decreasing order of the non-empty arguments ratio of the role, based on the intuition that more informative (non-empty) arguments in the path history can facilitate better subsequent argument identifications during the recurrent decoding, and we also validate this intuition by comparing with models trained on some randomly permuted role orders.

\begin{table*}[htb]\small
\centering
\begin{tabular}{ c | l | l}
\toprule
\textbf{Module} & \textbf{Hyper-parameter} & \textbf{Value} \\
\midrule
\multirow{3}{*}{\tabincell{c}{Input \\ Representation}}
& the maximum sentence number & $64$ \\
& the maximum sentence length & $128$ \\
& $d_w$ (the embedding size) & $768$ \\
\midrule
\multirow{2}{*}{\tabincell{c}{Entity \\ Recognition}}
& the tagging scheme & \texttt{BIO} (Begin, Inside, Other) \\
& the hidden size & $768$ (same to $d_w$) \\
\midrule
\multirow{3}{*}{\tabincell{c}{Transformer-1 \\ Transformer-2 \\ Transformer-3}}
& the number of layers & 4 \\
& the size of the hidden layer & $768$ (same to $d_w$) \\
& the size of the feed-forward layer & $1,024$ \\
\midrule
\multirow{12}{*}{\tabincell{c}{Optimization}}
& the optimizer & Adam \\
& the learning rate & $1e^{-4}$ \\
& the batch size & 64 (with 32 P40 GPUs) \\
& the training epoch & 100 \\
& the loss reduction type & sum \\
& $\lambda_1$ & 0.05 \\
& $\lambda_2, \lambda_3$ & 0.95 \\
& $\gamma$ & 3 \\
& the dropout probability & 0.1 \\
& the scheduled-sampling beginning & $10^{th}$ epoch \\
& the scheduled-sampling ending & $20^{th}$ epoch \\
& \tabincell{l}{the scheduled probability of employing \\ gold entity mentions} & \tabincell{l}{decreasing from $1.0$ to $0.1$ linearly \\ during the scheduled epochs} \\
\bottomrule
\end{tabular}
\caption{The hyper-parameter setting.}
\label{table:hyper_para}
\end{table*}

\subsection{Pseudo Codes for the EDAG Generation}
\label{sec:pseudo_codes}

We provide pseudo codes about how to calculate the EDAG loss for a given EDAG structure (Algorithm~\ref{alg:edga_train}) and how to generate an EDAG at inference (Algorithm~\ref{alg:edga_eval}) to facilitate better understanding of the EDAG generation.

\begin{algorithm*}[tb]
\SetAlgoLined
\KwIn{the EDAG structure $\bm{\mathcal{G}}$ for each triggered event, the entity tensor $\bm{e^d} = [\bm{e^d_1},\cdots,\bm{e^d_{N_e}}]$, the sentence tensor $\bm{c^d} = [\bm{c^d_1},\cdots,\bm{c^d_{N_s}}]$\;}
\KwOut{the EDAG generation loss $L_{dag}$\;}
Initialize the loss $L_{dag} = 0$\;
\For{each event type}{
  \If{is triggered}{
    Initialize the memory tensor of the virtual starting node as $\bm{c^d}$\;
    \For{each role type following the predefined order}{
      \For{each entity node of the last role in $\bm{\mathcal{G}}$}{
        Look up the memory tensor of this node as $\bm{m_t}$\;
        Get path-expanding labels from $\bm{\mathcal{G}}$\;
        Calculate the path-expanding classification loss $L_{pe}$\;
        Update the EDAG generation loss as $L_{dag} = L_{dag} + L_{pe}$\;
        
        \For{each entity $i$ known to be expanded in the current role}{
          Set the memory tensor for the corresponding entity node of the current role as $[\bm{m_t}, \bm{e^d_i}]$;
        }
      }
    }
  }
}
\caption{Pseudo codes to calculate the loss for the EDAG generation}
\label{alg:edga_train}
\end{algorithm*}

\begin{algorithm*}[tb]
\SetAlgoLined
\KwIn{the entity tensor $\bm{e^d} = [\bm{e^d_1},\cdots,\bm{e^d_{N_e}}]$, the sentence tensor $\bm{c^d} = [\bm{c^d_1},\cdots,\bm{c^d_{N_s}}]$\;}
\KwOut{the EDAG structure for each triggered event\;}
\For{each event type}{
  \If{is triggered}{
    Initialize the EDAG structure with a virtual starting node\;
    Initialize the memory tensor of the virtual starting node as $\bm{c^d}$\;
    \For{each role type following the predefined order}{
      \For{each leaf node of the current EDAG}{
        Look up the memory tensor of this node as $\bm{m_t}$\;
        Get path-expanding predictions\;
        
        \For{each entity $i$ predicted to be expanded in the current role}{
          Create a new node of entity $i$ for the current role\;
          Update the EDAG structure by connecting the leaf node to the new entity node\;
          Set the memory tensor of that new node as $[\bm{m_t}, \bm{e^d_i}]$;
        }
      }
    }
  }
}
\caption{Pseudo codes to generate an EDAG at inference}
\label{alg:edga_eval}
\end{algorithm*}

\subsection{Additional Evaluation Results}
\label{sec:add_eval_res}
In the main body, we show the end-to-end evaluation results of DEE for different models when facing \textit{arguments-scattering} and \textit{multi-event} challenges.
Here, Table~\ref{tab:add_res} complemented both evaluation results and corresponding analyses for entity extraction and event triggering, two preceding sub-tasks before filling the event table.

\begin{table*}[tb]\small
\centering
\addtolength{\tabcolsep}{-3.2pt}
\begin{tabular}{l | ccc | ccc | ccc | ccc | ccc | ccc }
\toprule
\multirow{2}{*}{\textbf{Model}}
& \multicolumn{3}{c|}{\tabincell{c}{\textbf{Entity} \\ \textbf{Extraction}}}
& \multicolumn{3}{c|}{\tabincell{c}{\textbf{EF} \\ \textbf{Triggering}}}
& \multicolumn{3}{c|}{\tabincell{c}{\textbf{ER} \\ \textbf{Triggering}}}
& \multicolumn{3}{c|}{\tabincell{c}{\textbf{EU} \\ \textbf{Triggering}}}
& \multicolumn{3}{c|}{\tabincell{c}{\textbf{EO} \\ \textbf{Triggering}}}
& \multicolumn{3}{c}{\tabincell{c}{\textbf{EP} \\ \textbf{Triggering}}}
\\
& \textbf{P.} & \textbf{R.} & \textbf{F1}
& \textbf{P.} & \textbf{R.} & \textbf{F1}
& \textbf{P.} & \textbf{R.} & \textbf{F1}
& \textbf{P.} & \textbf{R.} & \textbf{F1}
& \textbf{P.} & \textbf{R.} & \textbf{F1}
& \textbf{P.} & \textbf{R.} & \textbf{F1}
\\
\midrule
DCFEE-O & 87.7 & 90.8 & 89.3  & 99.3 & 72.1 & 83.5  & 100.0 & 90.7 & 95.1  & 95.5 & 68.1 & 79.5  & 97.0 & 66.2 & 78.7  & 98.7 & 88.7 & 93.4  
\\
DCFEE-M & 88.5 & 90.3 & 89.4  & 99.3 & 66.7 & 79.8  & 100.0 & 88.0 & 93.6  & 95.7 & 63.8 & 76.6  & 94.3 & 71.7 & 81.4  & 98.5 & 86.4 & 92.1  
\\
\midrule
GreedyDec & 89.0 & 89.7 & 89.3  & 100.0 & 98.5 & 99.3  & 100.0 & 99.8 & 99.9  & 97.2 & 98.2 & 97.7  & 99.1 & 95.1 & 97.1  & 99.0 & 99.8 & 99.4 
\\
Doc2EDAG & 89.0 & 89.6 & 89.3  & 100.0 & 99.5 & 99.8  & 100.0 & 99.6 & 99.8  & 97.9 & 97.9 & 97.9  & 97.3 & 94.8 & 96.0  & 99.4 & 100.0 & 99.7
\\
\bottomrule
\end{tabular}
\caption{Evaluation results of entity extraction and event triggering for each event type on the test set, where we can observe that 1) different models produce roughly consistent entity-extraction performance, which corresponds to our setting that all models share the same architecture when extracting entities; 2) the document-level event triggering is superior to the sentence-level event triggering (key-event sentence detection used in DCFEE), because both the DS-based labeling and the event-triggering learning can be more accurate and robust at the document level, and the assumption to identify key-event sentences used by DCFEE is hard to fit all event types well.}
\label{tab:add_res}
\end{table*}

\subsection{Case Studies}
\label{sec:case}
\begin{figure*}[tb]
\centering
\includegraphics[width=0.95\linewidth]{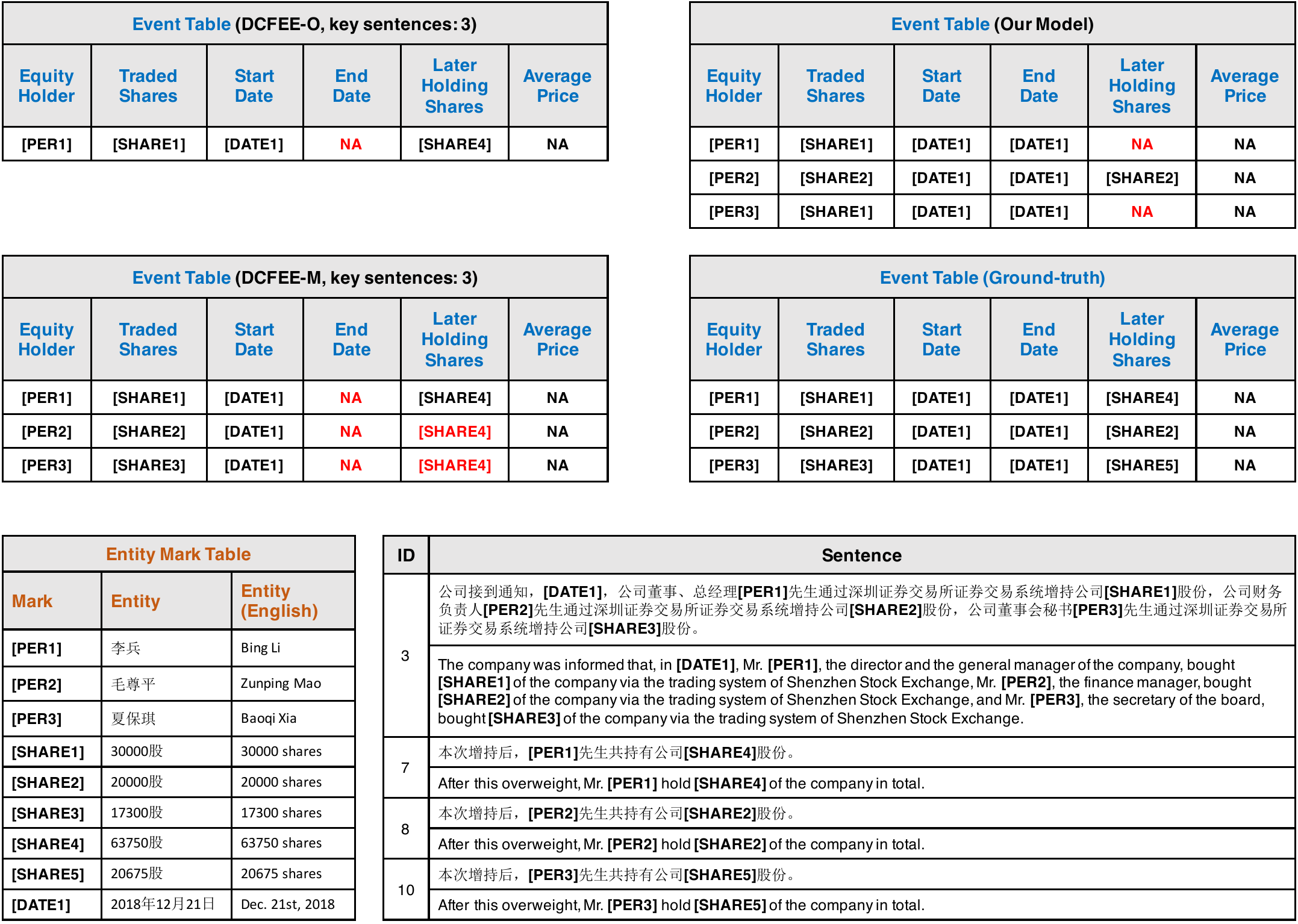}
\caption{In this case, there are three \textit{equity overweight} (EO) events mentioned by the documents.
Although DCFEE-O correctly identifies the key sentence (ID 3), it cannot decide how many events being expressed by this sentence, as its SEE module can only fulfill the sequence tagging task.
Therefore, we implement another version, DCFEE-M, which guess possible events by the position closeness, and indeed DCFEE-M produce multiple partially correct events in this case.
However, the arguments-completion stage of DCFEE-M is context-agnostic,
which is the reason that DCFEE-M does not produce correct arguments for the \textit{End Date} role (``DATE1'' is already assigned with the \textit{Start Date} role) and the \textit{Later Holding Shares} role (the closest valid entity is ``[SHARE4]'').
Moreover, though achieving better results for this case, DCFEE-M is inferior to DCFEE-O in terms of the whole test set (shown in the paper), since the naive multi-event guessing fails on many other cases, such as the case shown in Figure~\ref{fig:case_eu}.
Since our model can perform the end-to-end context-aware inference, it produces much better results than existing solutions, though it also misses two arguments for the \textit{Later Holding Shares} role.
After the careful examination, we find that the empty ratio of this role is pretty high during training, and thus our model prefers to be conservative in assigning entities with this role.
This case also implies that there still exists some promotion spaces for the Doc2EDAG paradigm.}
\label{fig:case_eo}
\end{figure*}

\begin{figure*}[tb]
\centering
\includegraphics[width=0.95\linewidth]{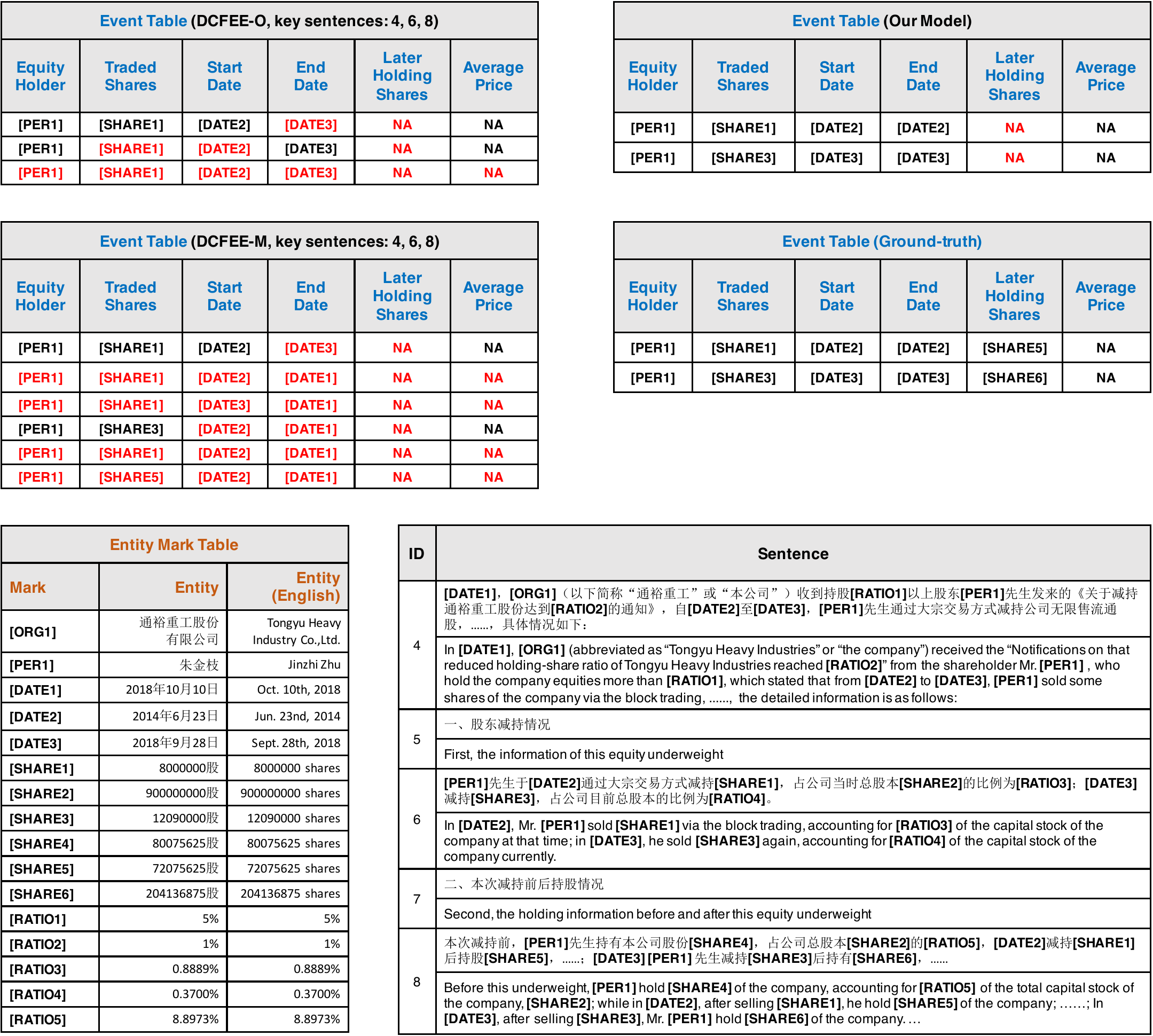}
\caption{This case shows the typical false positive errors made by DCFEE models.
Although the document only contains two distinct \textit{Equity Underweight} events in total, different sentences mention these events multiple times (ID 4, 6 and 8).
However, the key-sentence detection module of DCFEE models cannot differentiate duplicated event mentions elegantly.
Therefore, both of them produce duplicated event records.
Especially, DCFEE-M, guessing multiple event mentions from a single sentence, suffers severe false positive errors in this case.
In contrast, our model is naturally robust to such data characteristics, since we conduct the event table filling at the document level.
The only missing arguments, belong to the \textit{Later Holding Shares} role, are partially caused by the restriction of the maximum sentence length at the input stage (ID 8).
}
\label{fig:case_eu}
\end{figure*}

\begin{figure*}[tb]
\centering
\includegraphics[width=0.95\linewidth]{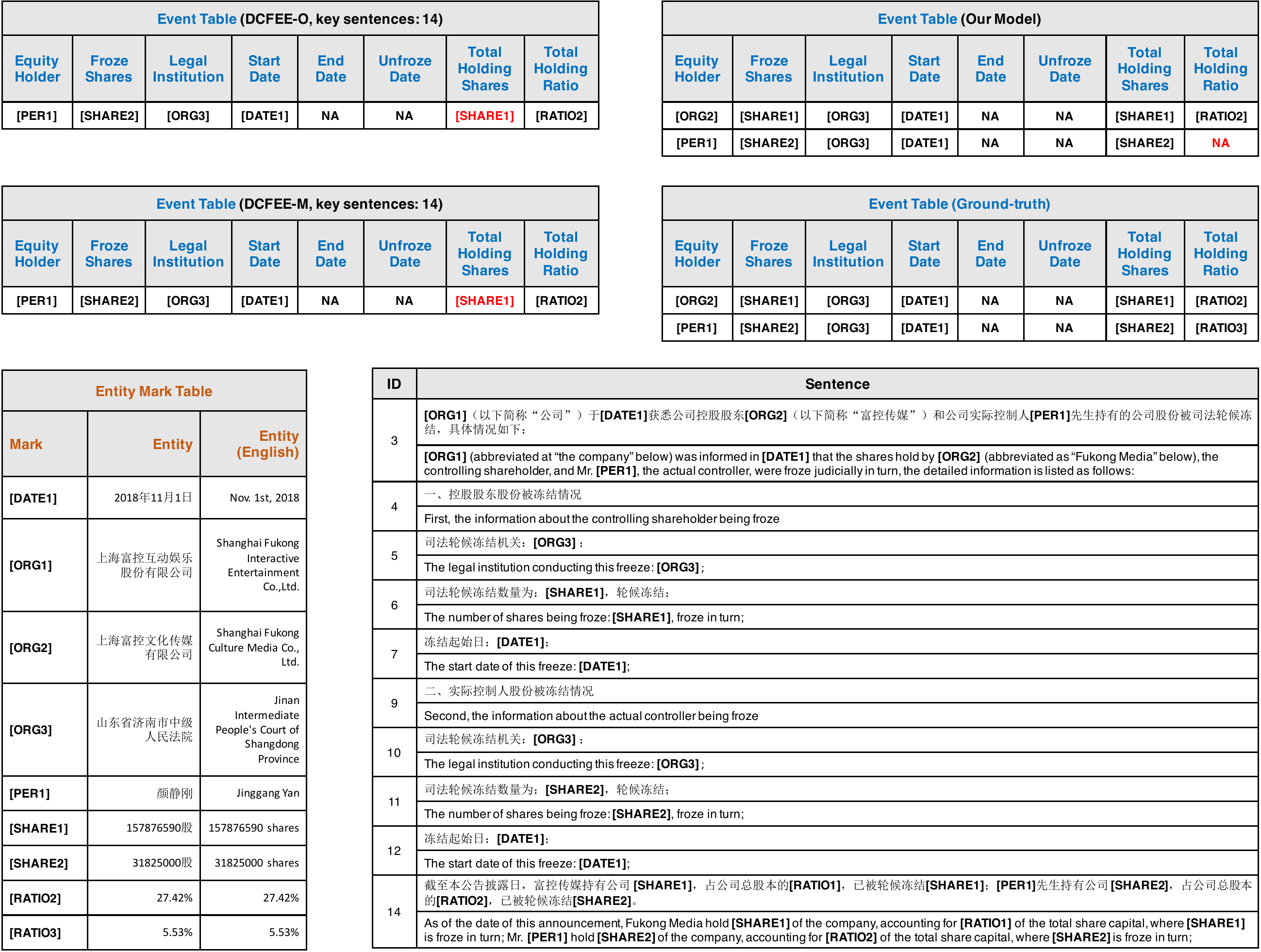}
\caption{
This case, containing two \textit{Equity Freeze} events, is a typical example that violates the key-sentence labeling assumption of DCFEE, which assumes the sentence containing the most arguments as the key-event one. 
We can observe that the core arguments of the event scatter across multiple sub-sentences, such as ID 5, 6, 7, 10, 11 and 12, 
but DCFEE-O and DCFEE-M treat the summary sentence (ID 14) as the key-event one.
However, the single sentence (ID 14) summarizes these two event records,
and DCFEE models cannot address such multi-event sentences elegantly.
Note that, each text snippet of ID 5, 6, 7, 10, 11 and 12 is not a complete sentence,
but these text snippets are presented in a list manner and some of them even do not have ending punctuations (ID 4, 9). 
We have tried to merge such short snippets into a single long sentence,
but applying this merge on the whole dataset can hurt the performance of DCFEE models on other event types.
Thus, we drop this preprocessing option.
In contrast, our model is immune to such merging and even benefit with a faster speed due to fewer sentences to be encoded.
In terms of the extraction performance, our model correctly identifies these two events and arranges entities into proper table columns with only one missing argument for the \textit{Total Holding Ratio} role.
While DCFEE models miss one event and inevitably make mistakes when completing missing arguments for the key-event sentence.
}
\label{fig:case_ef}
\end{figure*}

In addition to the \textit{Equity Pledge} example included by the paper,
we show another three cases in Figure~\ref{fig:case_eo},~\ref{fig:case_eu} and~\ref{fig:case_ef} for the \textit{Equity Overweight}, \textit{Equity Underweight} and \textit{Equity Freeze} events, respectively, to intuitively illustrate why Doc2EDAG, the truly end-to-end model, is better.
For all figures, we color the wrong predicted arguments as red and present detailed explanations.

\end{document}